\pdfoutput=1

\documentclass[11pt]{article}

\usepackage[]{ACL2023}

\usepackage{times}
\usepackage{latexsym}
\usepackage{graphicx}
\usepackage{amsmath}
\usepackage{amssymb}
\usepackage{times}
\usepackage{url}
\usepackage{latexsym}
\usepackage{caption} 
\usepackage{subcaption}
\usepackage{amsmath}
\usepackage{enumitem}
\usepackage{microtype}

\usepackage{xcolor}
\usepackage{booktabs, multirow} 
\usepackage{lipsum}
\usepackage{siunitx}
\usepackage{hyperref}
\usepackage{cleveref}
\usepackage[T1]{fontenc}

\usepackage[utf8]{inputenc}

\usepackage{microtype}

\usepackage{inconsolata}
\def\XXX#1{}

\title{Thesis proposal: Are We Losing Textual Diversity to Natural Language Processing?}

\author{Josef Jon \\
  Charles University, Faculty of Mathematics and Physics \\
Institute of Formal and Applied Linguistics \\
Prague, Czech Republic \\
  \texttt{jon@ufal.mff.cuni.cz}}

\begin{document}
\maketitle
\begin{abstract}
This thesis argues that the currently widely used Natural Language Processing algorithms possibly have various limitations related to the properties of the texts they handle and produce. With the wide adoption of these tools in rapid progress, we must ask what these limitations are and what are the possible implications of integrating such tools even more deeply into our daily lives.

As a testbed, we have chosen the task of Neural Machine Translation (NMT). Nevertheless, we aim for general insights and outcomes, applicable even to current Large Language Models (LLMs). We ask whether the algorithms used in NMT have inherent inductive biases that are beneficial for most types of inputs but might harm the processing of untypical texts.
To explore this hypothesis, we define a set of measures to quantify text diversity based on its statistical properties, like uniformity or rhythmicity of word-level surprisal, on multiple scales (sentence, discourse, language).  We then conduct a series of experiments to investigate whether NMT systems struggle with maintaining the diversity of such texts, potentially reducing the richness of the language generated by these systems, compared to human translators.

We search for potential causes of these limitations rooted in training objectives and decoding algorithms. Our ultimate goal is to develop alternatives that do not enforce uniformity in the distribution of statistical properties in the output and that allow for better global planning of the translation, taking into account the intrinsic ambiguity of the translation task. 
\end{abstract}

\section{Introduction}
Language technologies powered by machine learning, such as predictive typing, machine translation, text generation, or chat assistants have become deeply integrated into our daily lives. With the advancement of Large Language Models (LLMs), we can expect that interaction with such tools will become an even more integral part of our work and social interactions. Yet, many important questions regarding these tools remain unanswered.  

Do they understand language the same way we do? 
 Are they equipped to capture and reflect the full spectrum of a language's intricacies, innovations, and diversity as effectively as a human?
Are there any biases within the algorithms themselves that can be beneficial for processing ordinary types of texts, but harmful for specific cases that deviate from the usual rules found in mundane text content?

\begin{figure}
    \centering
    \includegraphics[width=1.0\linewidth]{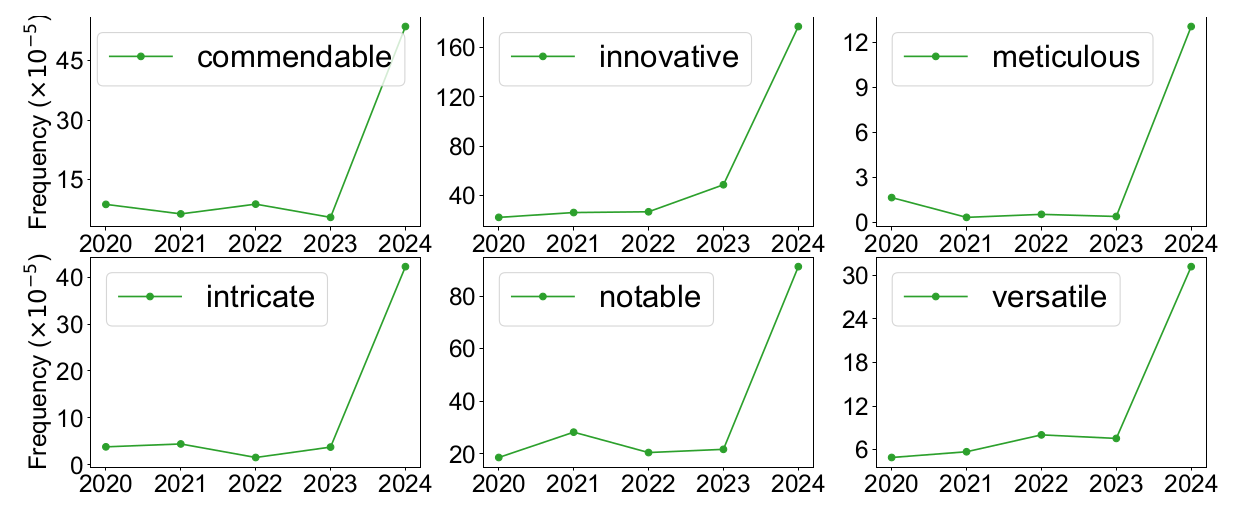}
    \caption{Usage of words that are often produced by ChatGPT in ICLR reviews in years 2020 to 2024. Increased frequency of these words in 2024 shows the prevalence of using ChatGPT for writing reviews. From \citet{liang2024monitoring}. }
    \label{fig:reviews}
\end{figure}

As our interaction with these machines deepens, is it possible that the users (i.e. most of the world's population) will adapt their language, simplify it for the machine to understand better? Could this  lead to further proliferation of simplistic, mundane text, which will be in turn used to train a new generation of models, making them even less adept at processing surprising inputs?  Will machine-produced, monotonous content dominate our informational space, causing authentic texts to become lost within it?\footnote{See Figure \ref{fig:reviews} for an example of such pollution already happening within the field of machine learning research itself.}

Ultimately, could this trend contribute to a loss of the diversity and richness of human language and, consequently, human thought?\footnote{For example, \citet{Kranich2014TranslationsAA} explains how language contact results in language changes}

This thesis explores such questions through the lens of one specific application of these language technologies: Neural Machine Translation (NMT). We investigate NMT's performance on texts that diverge from the norm not in terms of terminology or domain but in the structure and organization of their information content. 
While the challenge of domain diversity might be resolved by the field of Natural Language Processing (NLP) ``by itself", through the expansion of model and dataset sizes, we seek to uncover text types for which, regardless of model dimensions, the intrinsic properties are lost in translation. Our goal is to identify where NMT fails in preserving the essence of texts that defy conventional patterns, regardless of the amount of training data or model size. 

Even if we answered all the proposed questions positively, identifying a multitude of failure cases and adverse impacts of current tools, it would not change the rate at which these technologies are adopted. Thus, the ultimate outcome of this thesis is to propose a solution in the form of alternative decoding algorithms and training objectives, that allow the models to handle diverse texts properly.

\section{Problems and proposed solutions}
We briefly summarize the goals of this work in this section. The literature describing the characteristics we discuss here is listed in Section \ref{sec:related_work}. 

\subsection{Measuring diversity}

First, we establish criteria to assess the uniformity and typicality of the distribution of information on multiple scales, including the sentence level, the discourse level, and the whole structure of the language itself. Our methodology is based on the existing literature (described in Section \ref{sec:related_work}), and our objective is to associate the measures at each level with observable real-world phenomena.

At the word level, our exploration into measuring information content -- or surprisal -- examines its association with word-level reading times. We use surprisal estimates obtained by language models to predict the reading times. For sentences and paragraphs, we assess the uniformity of these surprisal values.  We evaluate different methodologies based on their correlation and their predictive capacity for sentence-level reading times and linguistic acceptability as judged by humans.

At the discourse level, we intend to observe rhythmic patterns in surprisal distribution to gauge engagement by predicting at what point the audience may stop reading an article or listening to a podcast. We hypothesize that a periodical change between quick and slow pace of information transmission has a large effect on the reader's enjoyment of the text.

Finally, in the context of entire languages, we will estimate the optimal information rate or channel capacity language-wide, including comparisons across different languages. 

\begin{figure}
    \centering
    \includegraphics[width=0.93\linewidth]{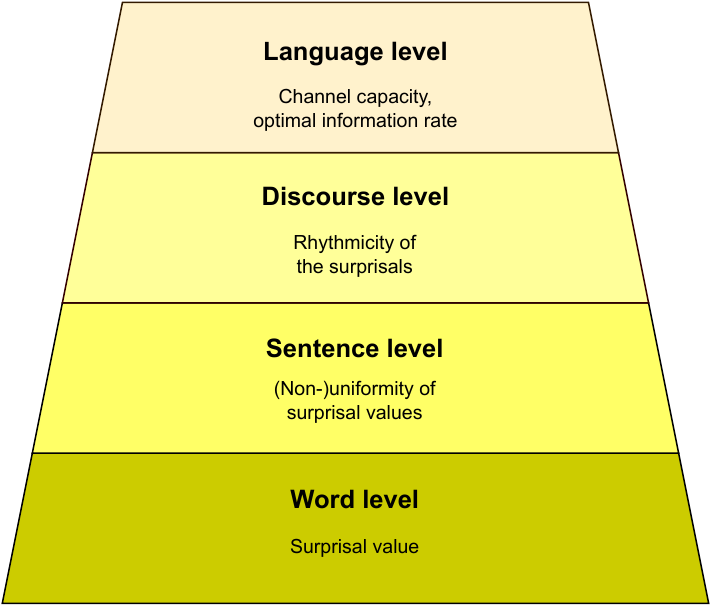}
    \caption{Proposed text properties that we plan to observe during the MT process. }
    \label{fig:properties}
\end{figure}

The properties we intend to observe during the translation process at different levels of the language are also listed in Figure \ref{fig:properties}.

Developing the measures for these properties will allow us to:
\begin{enumerate}[label={\arabic*)}]
    \item  Identify potentially problematic or difficult types of texts for NMT,
\item evaluate the current algorithms and our innovations,
\item  identify texts that were already translated by MT in the wild.
\end{enumerate}
Points 1) and 2) are integral to this thesis, and point 3) helps to counter one of the challenges that is already present in MT and arises more with the continuing adoption of LLMs: new models training on the outputs of previous generations of models due to pollution of the Web by machine-generated text. Such a self-feedback loop inevitably leads to loss of diversity over multiple iterations, up to a completely degenerated output~\citep{guo2023curious,briesch2023large}.

Section \ref{ssec:uid_measures} discusses our progress in this task.

\subsection{Diversity in MT}
We employ the uniformity measures explored in the previous part to answer one of the initial questions: \textit{Does NMT make the surprisal distribution in the translation more uniform than a human translator would?} We gather a diverse set of datasets to look for correlations in  surprisal distribution between source and both human and machine translation. We also compare the absolute values of the measures between professional human translation and MT. As we discuss in Section \ref{ssec:uid_mt}, the results we have gathered so far are not conclusive, as many factors play a role in computing the surprisal. However, we did find some evidence of MT failing to maintain non-uniformity of surprisal distribution in highly surprising texts, like poetry. 

\subsection{Causes of excessive uniformity}
Assuming we will, in fact, find that the NMT struggles to produce non-uniform text, we will investigate the causes of this behavior. The most obvious culprit seems to be beam search decoding. The beam search was introduced as an approximation of the maximum-a-posteriori (MAP) decoding, which is in practice intractable. However, it has been shown, quite surprisingly, that it produces better results than the exact MAP decoding \citep{stahlberg-byrne-2019-nmt}, suggesting that it introduces some own inductive bias into the translation process. \citet{meister-etal-2020-beam} show that this bias is linked to uniformity -- beam search prefers hypotheses where the surprisal is uniformly distributed and the same results can be obtained with exact MAP decoding with uniformity regularizer.

We hypothesize that this regularization is necessary to produce high-quality MT outputs because of insufficiencies in the modeling.
The model is predominantly trained to generate the next word given the source text and the previously produced target text, a process that does not inherently encourage global planning strategies. 
Without beam search enforcing uniformity of the surprisals, the model would make surprising decisions that it could in theory "balance out" in the future, but, given the lack of global planning, it performs this long-range balancing poorly. Beam search masks this inadequacy of the model by enforcing surprisal uniformity. 

This leads us to a conclusion: To replace beam search in order to improve the translation of non-uniform texts, we should also look into training objectives and improve the global planning capabilities of the model.

\subsection{Alternative decoding algorithm}
We propose an alternative decoding algorithm for NMT. Sampling-based algorithms are commonly used in Large Language Models (LLMs), but they have been shown unfit for traditional NMT with smaller model sizes. However, Minimum Bayes risk decoding (MBR) with sampled hypotheses has been very successful recently and we extend this approach by using a genetic algorithm to combine and modify the translation candidates \citep{jon-bojar-2023-breeding}, see Section \ref{ssec:decoding}. We show that it can be successfully used to improve translation quality \citep{jon-etal-2023-cuni} or find biases and flaws in MT evaluation metrics \citep{jon2024GAATME}.
We are circumventing the problem of too uniform surprisals by using MT quality evaluation and estimation metrics to guide the decoding, ignoring the probability estimated by the model.

In the future, we will also address decoding in LLMs and compare the properties of LLM-generated text to the conventional NMT models.

\begin{figure*}[ht]
\vspace{-40px}
    \centering
    \includegraphics[width=0.96\textwidth]{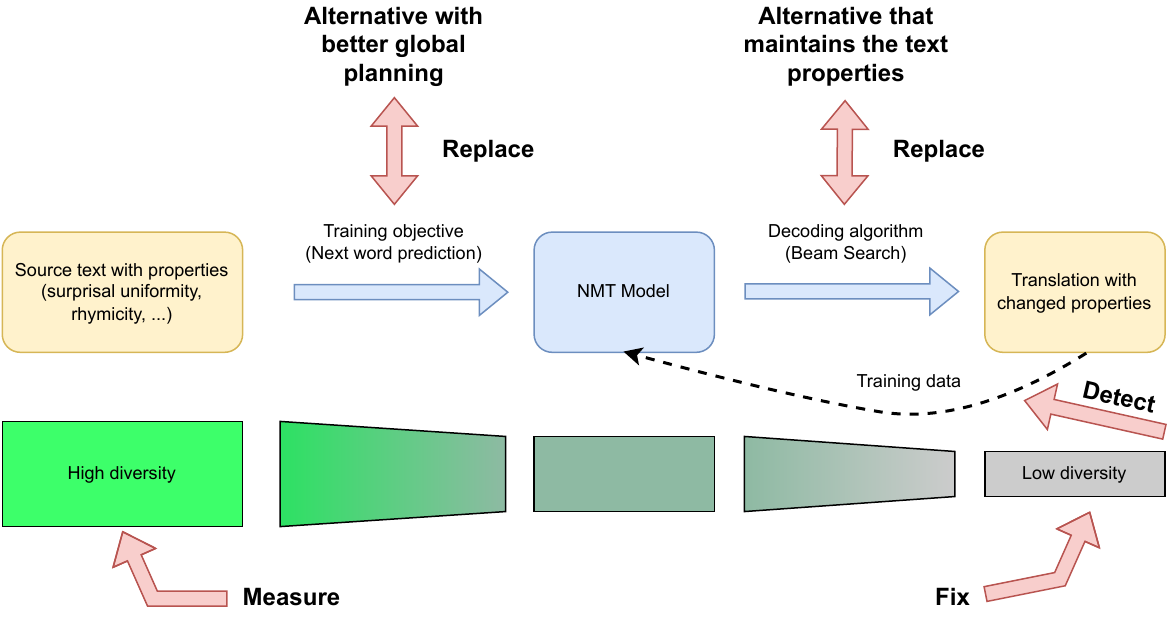}
    \caption{Illustration of the current NMT process and the points addressed in this thesis (red arrows).}
    \label{fig:problem}
\end{figure*}

\subsection{Alternative training objective}
\label{ssec:alternative_objective}
In line with the previous reasoning, we have also investigated alternative training objective functions for NMT to allow for better global planning of the translation. 
The intuitive reasoning behind this is that global planning allows the model to estimate how much probability the best solution has from the start, so it can plan for more uneven distribution of it throughout the output sequence, taking more ``risks" and selecting higher surprisal tokens, since it knows it will get the probability ``back" in the future timesteps. We assume that beam search negates this behavior, which is beneficial for models where the global planning contains a lot of mistakes.
 
 The approaches can range from simply predicting more future tokens at a single timestep \citep{pal-etal-2023-future,stern2018blockwise}, through sentence-level objectives \citep{goyal-etal-2019-empirical,lu-etal-2020-mixed}, to modeling hierarchical structures of the text \citep{ainslie-etal-2020-etc}.
 
The long-term planning abilities are also inherently related to the intrinsic uncertainty of the translation task, stemming from the fact that each source sentence has multiple correct translations. Current modeling approaches are inadequate since they only model a single probability distribution over the target sentence (although viable alternatives exist; \citet{stahlberg-kumar-2022-jam}). In left-to-right decoding, this uncertainty rises with the distance of the future we are trying to predict, as at each next token, the word choices are split into many possible, and all correct, pathways. Thus, the inadequacy of the modeling becomes even more pressing in our distant future modeling use case.

We propose using an auxiliary training objective, Contrastive Predictive Coding (CPC; \citet{oord2019representation}), that forces the internal representations of the model at each timestep to be more similar to future internal representations on multiple levels of hierarchy (words, phrases, sentences, and paragraphs). This allows us to model the future, while simultaneously alleviating the uncertainty issue, since instead of forcing a prediction of a single token, we predict an internal representation of a more abstract nature at some point in the future. However, in current models, the internal representations are still only a single-point estimates of the corresponding words or other linguistic phenomena. We will explore approaches to model the uncertainty in embeddings themselves, similar to \citet{Kesiraju_2020}.

Conversely, the past context is also modeled inadequately in NMT. 
Traditional \textit{teacher forcing} practice, which feeds the previous reference token into the decoder during training, imposes a single ``correct'' past context on the model. This causes \textit{exposure bias}: the model never sees its own generated prefixes during the training, but they are used at test time \citep{ranzato2016sequence}. This bias negatively affects the fidelity of the statistical properties of generated text to the training data. In a way, this discourages the model from planning into the future (or accounting for possible futures) when generating the current token, since in the next step, the current token is thrown away anyway.  
To address this, we will look into alternatives that, for instance, introduce a lattice of possible synonyms rather than a single reference token, offering a range of potential past contexts, or we will alternate between model's predictions and gold tokens during the training.
Additionally, we aim to explore sentence-level training objectives that permit the model to generate complete hypotheses freely, evaluating them semantically against the reference, inspiring ourselves in modern MT quality estimation approaches.

Reinforcement learning from human feedback (RLHF; \citet{stiennon2022learning,ouyang2022training}) has recently gained popularity with LLMs and it can be also viewed as a sentence-level objective. Later in our work, we will explore it along with other techniques used in LLMs.



See Section \ref{ssec:cpc} for details of the work on training objectives carried out so far.

\subsection{Objectives}
Based on the previous observations, we set out to reach 5 objectives:
\begin{itemize}
    \item \textbf{O1:} Identify texts that are inherently challenging for today's NMT systems, create a simple measure to score the non-uniformity of the surprisal distribution on multiple levels.
    \item \textbf{O2:} Identify the root causes of the problems: How do different steps of the NMT process affect the properties of the output?
    \item \textbf{O3:} Develop alternative decoding algorithms that mitigate the issues.
    \item \textbf{O4:} Develop alternative training objectives that allow for better global planning.
    \item \textbf{O5:} Perform a case study where the resulting system produces better translations in a real-world scenario.
\end{itemize}

We illustrate the NMT process, the potential problems we posit, and the places where we intervene to counter them (red arrows) in Figure \ref{fig:problem}.

\section{Characteristics of NMT-produced language}
\label{sec:related_work}
In this section, we summarize the related work on properties of translations produced by NMT and the origins of these properties.

\subsection{Lexical diversity in NMT}
Lexical diversity is a measure of the richness of a text's vocabulary and common metrics include the Type-Token Ratio (TTR) and the Measure of Textual Lexical Diversity (MTLD; \citealp{mccarthy2005assessment}). Studies have shown discrepancies in lexical diversity between NMT and human translations, often highlighting a reduction in lexical richness and the reinforcement of biases within NMT systems.

\citet{vanmassenhove-etal-2019-lost}  found that both phrase-based MT and NMT systems overgeneralize, leading to a loss of lexical richness by generating frequent words more often and rare words less often than in their training data. The authors didn't use subword segmentation, which is one of the main tools for NMT to generate morphologically rich outputs, thus the results are not applicable to fully-fledged NMT systems. They addressed these concerns in their follow-up work, reaching similar conclusions \citep{vanmassenhove-etal-2021-machine}.

\citet{toral-2019-post} found that both raw and post-edited machine translations (MT) offer less lexical diversity than human translations, exhibiting issues like simplification and source language interference. 

Junczys-Dowmunt\footnote{\url{https://marian-nmt.github.io/2020/01/22/lexical-diversity.html}} shows that NMT systems evaluated in the WMT19 News task matched human lexical diversity. Nonetheless, the domain of this task -- newspaper articles -- might be an inadequate benchmark for evaluating lexical diversity in NMT, since extensive lexical creativity is usually not required for this type of text.
\citet{Brglez2022Lexical} initially found higher lexical diversity in English-Slovenian NMT than in human translations. Their analysis clarified that perceived diversity often stemmed from translation errors and inconsistencies. This finding underscores an important insight we also address in our experiments: measures of creativity or diversity in MT must always be balanced with considerations of translation quality. Otherwise, translations with errors might be mischaracterized as innovative or diverse, even if they are simply wrong.

\subsection{Skewed word frequencies}
The root cause of the lexical uniformity is that the distribution of word frequencies in NMT outputs is skewed towards high-probability words.
This leads to more common words being used even more frequently in translations, while rare words become even scarcer \citep{ott2018analyzing, koehn-knowles-2017-six}. 
Achieving a distribution of output words that mirrors the training data closely can best be attempted through sampling from the model's distribution. However, this approach often results in sub-optimal translation quality. 

To increase the translation quality of sampling-based approaches, sampling can be combined with Minimum Bayes Risk (MBR) decoding and advanced MT evaluation metrics. On the other hand, MBR still brings about a resurgence of the original bias, over-representing frequent words. This is because MBR selects hypotheses that bear the greatest similarity to other sampled hypotheses, disadvantaging rare words due to their infrequent sampling and consequent lower scores \citep{muller-sennrich-2021-understanding}. 

\subsection{Decoding Algorithms}

Decoding algorithms are the mechanism through which a subsequent word $y(t)$ is predicted in an output sequence at time step $t$, based on a probability distribution over the target vocabulary, conditioned upon the source sentence $x$ and the sequence of previously generated target words, formalized as $P(y(t) | x, y(t-1), \ldots, y(0))$. 
The probability of a whole target sentence is factorized by the chain rule as a product of these word-level probabilities. The Maximum A Posteriori (MAP) decoding is traditionally employed to identify the most probable translation under the model's distribution. However, the search space for the decoding is very large: the size of the target vocabulary to the power of the maximum sentence length (theoretically unbounded). Therefore, an approximation of MAP is used, most commonly the beam search \citep{och-ney-2004-alignment,graves2012sequence}. Both beam search and MAP in general have many shortcomings which are discussed in recent literature.

\citet{stahlberg-byrne-2019-nmt} introduce an exact decoding algorithm utilizing depth-first search that can be executed within reasonable time constraints for most sentences. Despite achieving optimal model probabilities, a significant portion of the translations produced end up as empty strings. Adjustments like length normalization improved this but still produced translations of lower quality than those from beam search, measured by BLEU score. 
 This discrepancy suggests two conclusions. First, there might be an issue with how the whole translation task is modeled, since often the global optimum of these probabilistic models is an empty string.
 It also suggests beam search's effectiveness may not be due to its accuracy as a MAP approximation, but rather due to its ability to overcome flaws in the probabilistic modeling approach, challenging the reliance on the model's probability as the selection criterion.

 \citet{meister-etal-2020-beam} also explore the reasons why beam search produces higher quality outputs than an exact MAP search. They propose that the beam search itself introduces an inductive bias.  To find it, they reverse engineer the objective that beam search is a solution for, i.e. they try to build MAP-based algorithm that produces the same results as beam search. Their conclusion is that the exact MAP with a uniformity regularizer which enforces Uniform Information Distribution (UID) \citep{aylett2004,fenk1980,levy2006,bell2003,genzel-charniak-2002-entropy} behaves the same as beam search. The UID hypothesis posits a preference among language users for utterances that distribute information evenly.  In other words, beam search does not only look for the most probable solution but also prefers solutions where the probability is distributed evenly across the whole sentence. This property effectively conceals mistakes in NMT modeling and allows for the production of usable translations, even being dubbed the \textit{beam search blessing} by \citet{meister-etal-2020-beam}. \citet{wei-etal-2021-cognitive} employ a similar regularizer in the training of the model, which led to improved translation quality.


Beam search's preference for selecting hypotheses that adhere to UID principles at the sentence level suggests a potential conflict for certain text types, particularly those where an element of surprise or non-uniform information distribution is desired. This line of thinking forms the basis for the examination of alternative decoding algorithms within our study.

Work on pathologies of NMT, like \citet{koehn-knowles-2017-six} and \citet{stahlberg-byrne-2019-nmt}, sparked a debate on whether the standard training objective for machine translation, word-level maximum likelihood estimation, is suitable for modeling the task, and if the issues arise from the definition of the objective itself.

\citet{eikema-aziz-2020-map} suggest that the objective function is correct and the issue lies within the MAP decoding. They argue that the mode of the model’s distribution is not an adequate decision rule for as high-dimensional problem as NMT -– only a very small probability mass is given to the single most probable translation. 
Their analysis shows that sampling-based approaches produce translations more reflective of the training data's statistical properties, although they do not achieve optimal translation quality. They advocate for minimum Bayes risk (MBR; \citet{Goel2000MinimumBA}) decoding over beam search, as it covers a broader probability range and better preserves the training data's statistical properties.

MBR does not aim to identify translations with the highest model probability. Instead, it seeks translations that maximize a chosen utility function, often an MT evaluation metric such as BLEU or COMET. 
Notably, MBR does not require reference translations; the utility function is evaluated against the entire set of all possible translations of the input.

Again, the complete computation is intractable in practice, so sampling-based approximations are used, such as sampling a pool of hypotheses from the model and using them as both candidate translations and references, computing the utility function of each hypothesis with respect to all the others. 
The same authors discuss strategies to construct the pool of hypotheses in \citet{eikema-aziz-2022-sampling}.

MBR started to gain more popularity recently because of advances in MT metrics which can be used as the utility function \citep{amrhein-sennrich-2022-identifying,freitag-etal-2022-high,fernandes-etal-2022-quality,jon-etal-2022-cuni,jon-bojar-2023-breeding}. 

\subsection{Training objective}
Our discussion so far has already touched upon the appropriateness of the prevailing objective in NMT. 

Conventionally, this objective models the translation process as computing the probability of a target translation $y$ given a source sentence $x$, represented by a single probability distribution $P(y|x)$. 
 In practice, this is implemented by a softmax layer as the last layer of the decoder, which computes probability distribution over target vocabulary in each decoding timestep.
 Such a model implicitly assumes the existence of a single ``correct'' translation for every source sentence -- an assumption that contrasts with the linguistic reality where an extensive amount of valid translations can exist for a single source text. This formulation of the objective forces these valid translations to compete for representation within a single probability distribution,
 This does not allow the model to distinguish between two types of uncertainty: extrinsic uncertainty, caused by noisy training data, and intrinsic uncertainty, caused by the existence of multiple valid translations of a single source sentence. \citet{stahlberg-etal-2022-uncertainty} identify the intrinsic uncertainty as the main reason behind multiple pathologies in NMT, including the already discussed issues with MAP decoding.

SCONES (Single-label Contrastive Objective for Non-Exclusive Sequences; \citet{stahlberg-kumar-2022-jam}) aims to remove this single correct translation assumption by modeling the translation probabilities separately for each (source sentence, possible translation) pair, so that multiple valid translations from training data can be considered correct at the same time. 
The results suggest that using SCONES improves translation quality over many language pairs and it alleviates the problems that arise with MAP decoding described earlier – the inadequacy of mode and shifting of the text statistics compared to training data. 
The related work that we directly base our approach on is described in  \ref{ssec:alternative_objective}.
\section{Progress so far}
Here we describe the work towards the defined objectives that we have carried out so far.

\subsection{O1: Information distribution in language}
\label{ssec:uid_measures}
Surprisal theory, introduced by \citet{hale-2001-probabilistic}, relates cognitive effort to the surprisal value of words, suggesting that the effort to comprehend a word increases with its unpredictability given the context. The surprisal of an element $u_n$ in an utterance $\mathbf{u}$ is defined as $s(u_n) = -\log p(u_n|\mathbf{u}_{<n})$. The theory \citep{hale-2001-probabilistic} proposes that cognitive effort is proportional to surprisal: $$\mathrm{Effort}(u_n) \propto s(u_n)$$

Choosing reading time and linguistic acceptability ratings as proxies for the effort, we compared surprisal estimates computed by multiple language models across various datasets \citep{smith2013effect,futrell-etal-2018-natural,warstadt-etal-2019-neural} and we successfully confirmed this hypothesis on the word-level, as did many works before us  \citep{fernandez-monsalve-etal-2012-lexical,goodkind-bicknell-2018-predictive,oh2022does,oh-schuler-2023-surprisal}. 

Applying this concept to a longer sequence, like a sentence, leads to an unexpected conclusion: If we calculate the sentence's surprisal as the sum of surprisals of its individual words, and if this aggregate surprisal predicts the effort needed to process the sentence, then any way of distributing the information across the utterance is the same in terms of the effort needed for comprehension.

The Uniform Information Density (UID) theory, proposed by \citet{levy2006}, accounts for this unintuitive conclusion by proposing a super-linear relationship between surprisal levels and cognitive effort, factoring in utterance length  $N$,  suggesting sentences with evenly distributed surprisal are easier to comprehend:
 $$ \operatorname{Effort}(\mathbf{u}) \propto \sum_{n=1}^N s\left(u_n\right)^k+c \cdot N, k>1 $$


To illustrate the intuitive concept of surprisal uniformity, consider these two sentences: \begin{enumerate}[label=\bf{\Alph*)},leftmargin=0.6cm]
\item \textbf{More uniform:} \textit{"When she got home after a long day at work, she decided to relax by reading her favorite novel and having a cup of tea."}
\item \textbf{Less uniform:} \textit{"London's annual festival was filled with activities, food stands, windsurfing, and drinks, but the sudden unveiling of a Yetti statue caught everyone's attention."}
\end{enumerate}

\begin{figure}
\includegraphics[width=0.99\linewidth]{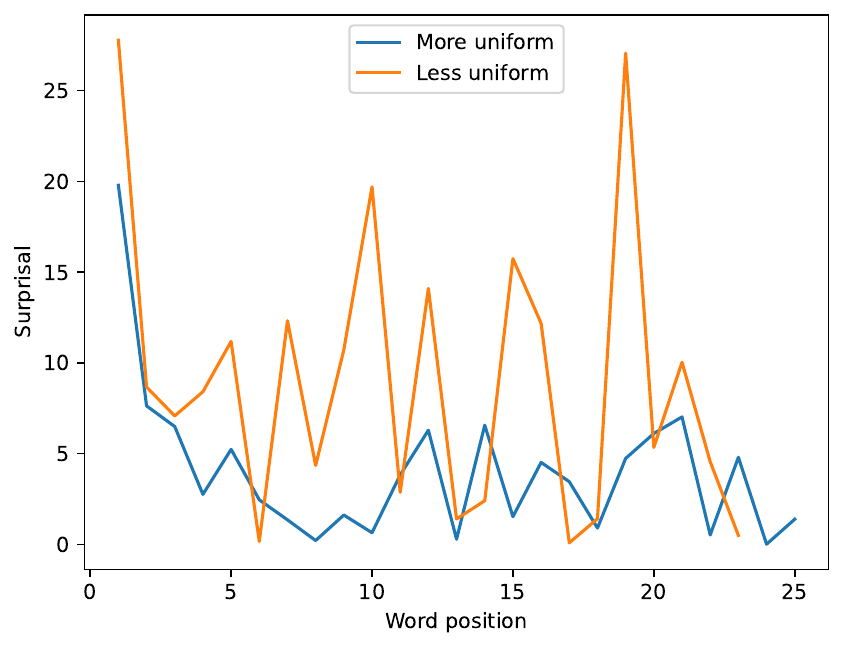}
\caption{Surprisal behavior for the two examples sentences, measured by GPT-2 model.}
\label{fig:surp_comparison}
\end{figure}

Most people would consider the second sentence as more surprising, as some of the words feel unexpected. We show the surprisal profiles of both sentences in Figure \ref{fig:surp_comparison}. Indeed, we can see that the profile of the second sentence (orange) looks less uniform. 

To operationalize our concept of uniformity and anchor it into the real world, we carried out assessments of various surprisal distribution uniformity measures for their correlation with human acceptability ratings and reading times, extending the work of \citet{meister-etal-2021-revisiting}. We explore measures like Local Variance (LV), Coefficient of Variation (CV), Global Variance (GV), Gini coefficient, and Super-linear Relationship (SL), and super-linear syntactic log-odds ratio (SLOR, \citealp{kann-etal-2018-sentence,pauls-klein-2012-large}):
\begin{itemize}
    \item $\operatorname{LV}(\mathbf{u})=\frac{1}{N-1} \sum_{n=2}^N\left(s\left(u_n\right)-s\left(u_{n-1}\right)\right)^2$
    \item  $\operatorname{CV}(\mathbf{u})=\frac{\sigma(\mathbf{u})}{\mu(\mathbf{u})}$
    \item $ \mathrm{GV}(\mathbf{u})=\frac{1}{N} \sum_{n=1}^N\left(s\left(u_n\right)-\mu(corpus)\right)^2 $
\item $ \operatorname{SL}(\mathbf{u})=\frac{1}{N} \sum_{n=1}^N s\left(u_n\right)^k \quad(k>1) $  
\item $ \operatorname{SLOR}(\mathbf{u})=\frac{1}{N} \sum_{n=1}^N s\left(u_n\right)^k - s_u\left(u_n\right)^k 
 \quad(k>1) $  

\end{itemize}
Function $s$ denotes surprisal in of a word in context, $s_n$ is a unigram, context-free surprisal.
We aimed to predict a sentence's linguistic acceptability and reading times using these surprisal distribution uniformity measures, employing statistical and machine learning methods like Pearson's r, linear regression (LR), Support Vector Machines (SVM), Multi-layer perceptron (MLP), Generalized linear models (GLM) and Linear mixed-effect models (LME). We sought to mostly replicate the results of \citet{meister-etal-2021-revisiting} and to add novel evaluation methods. We succeeded in obtaining similar results under the exact same conditions, i.e. some evidence to support the existence of a super-linear (SL) relationship. However, we have also learned that the outcome is very sensitive to methodological choices, like the dataset, language model used to calculate the surprisal, preprocessing and filtering choices and evaluation methods. In fact, the SL relationship for sentence-level reading times was refuted by the most recent and most extensive study so far \citep{rt_log}. This study presents strong evidence in favor of a simple linear relationship between word-level surprisals and effort (the team also included the authors of \citet{meister-etal-2021-revisiting}).

\begin{figure*}[ht]%
\vspace{-30px}
\centering
    \subfloat[\centering Pearson's r]{{\includegraphics[width=4.7cm]{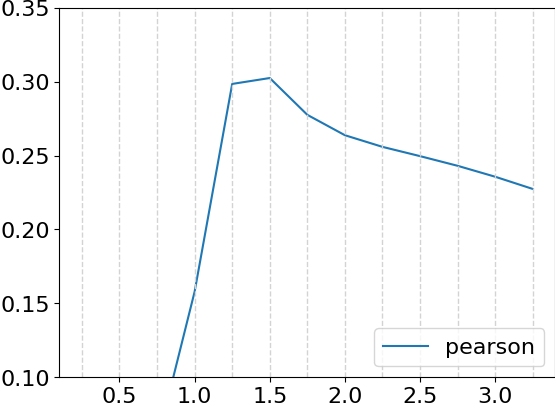}}}%
    \quad
        \subfloat[\centering Linear regression]{{\includegraphics[width=4.7cm]{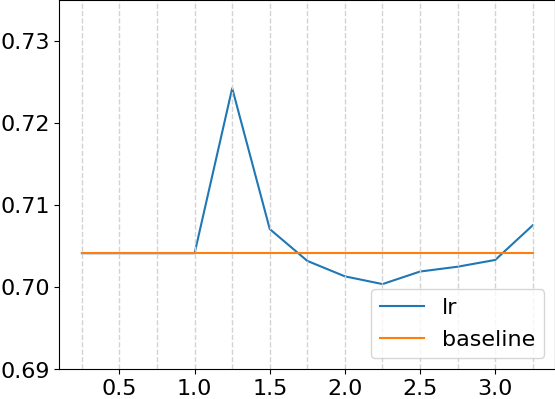} }}%
    \quad
    \subfloat[\centering GLM ]{{\includegraphics[width=5.2cm]{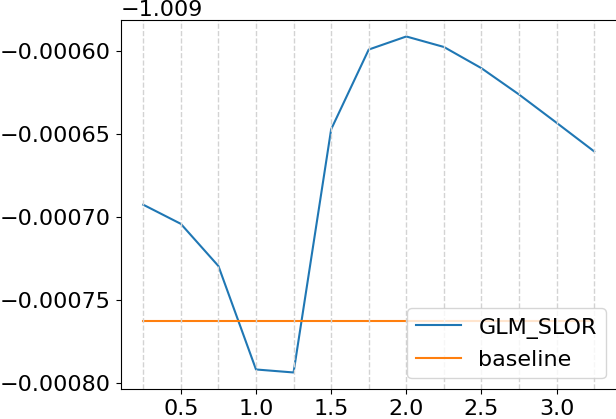} }}%

    \caption{Behavior of SLOR measure, depending on power $k$ used in the calculation.}%
    \label{fig:slor}%
\end{figure*}

We present the results for linguistic acceptability for the SLOR measure in Figure
\ref{fig:slor}.
We see that the measure correlates better with acceptability ratings for $k>1$, and it is slightly more predictive in LR and GLM models. According to SLOR, there might be some preference for sentences with more uniform surprisal distribution. 

\subsection{O1 and O2: MT and uniformity of surprisal}
\label{ssec:uid_mt}

\begin{figure*}[ht]%
\centering
    \subfloat[\centering Books]{{\includegraphics[width=7.5cm]{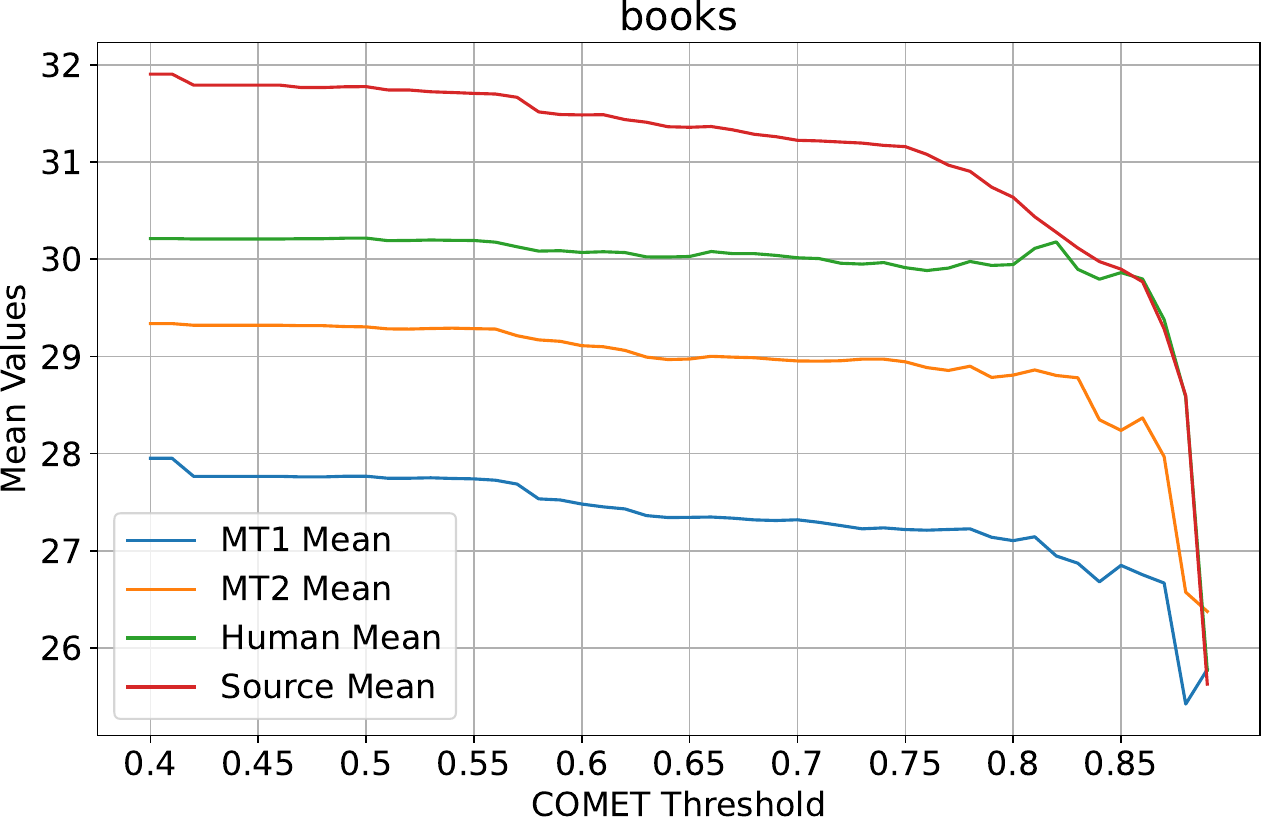}}}\quad%
    \subfloat[\centering Global]{{\includegraphics[width=7.5cm]{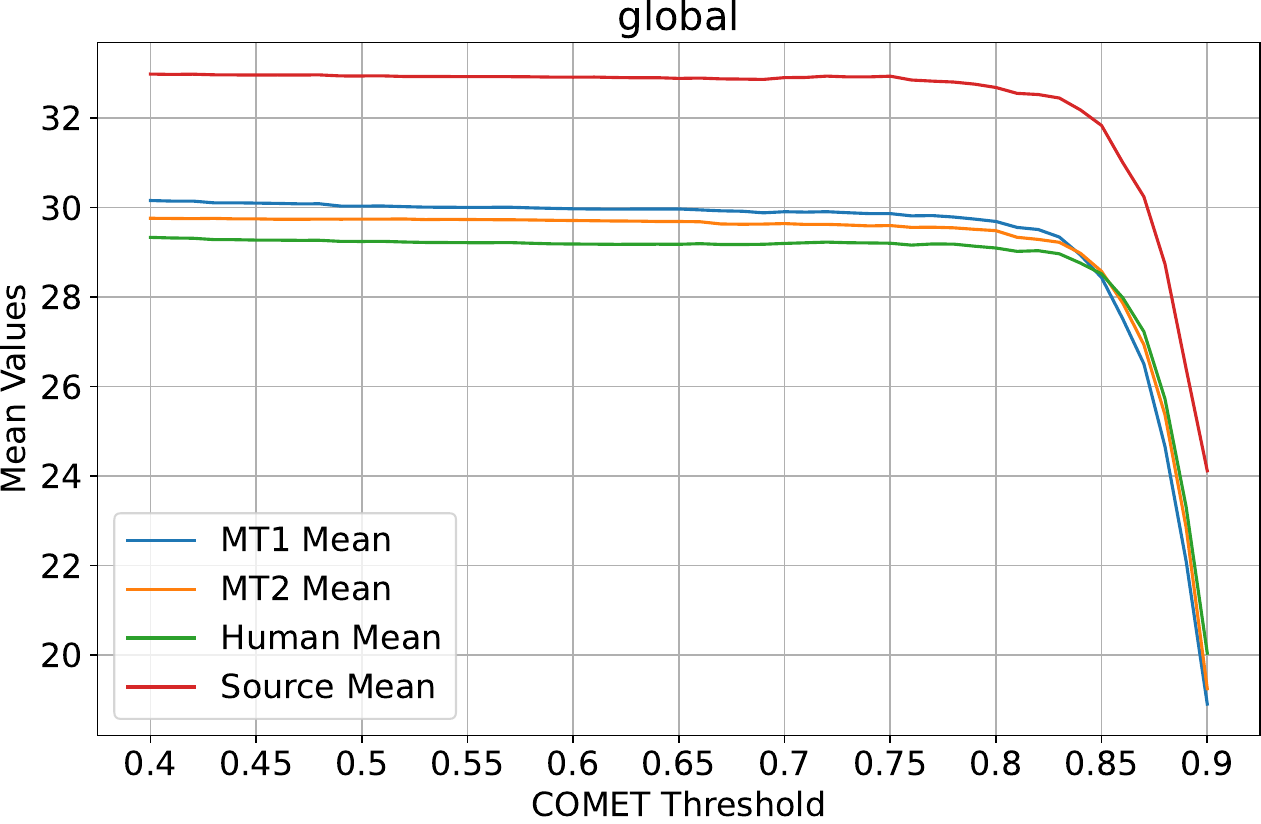} }}
    
    \subfloat[\centering WMT]{{\includegraphics[width=7.5cm]{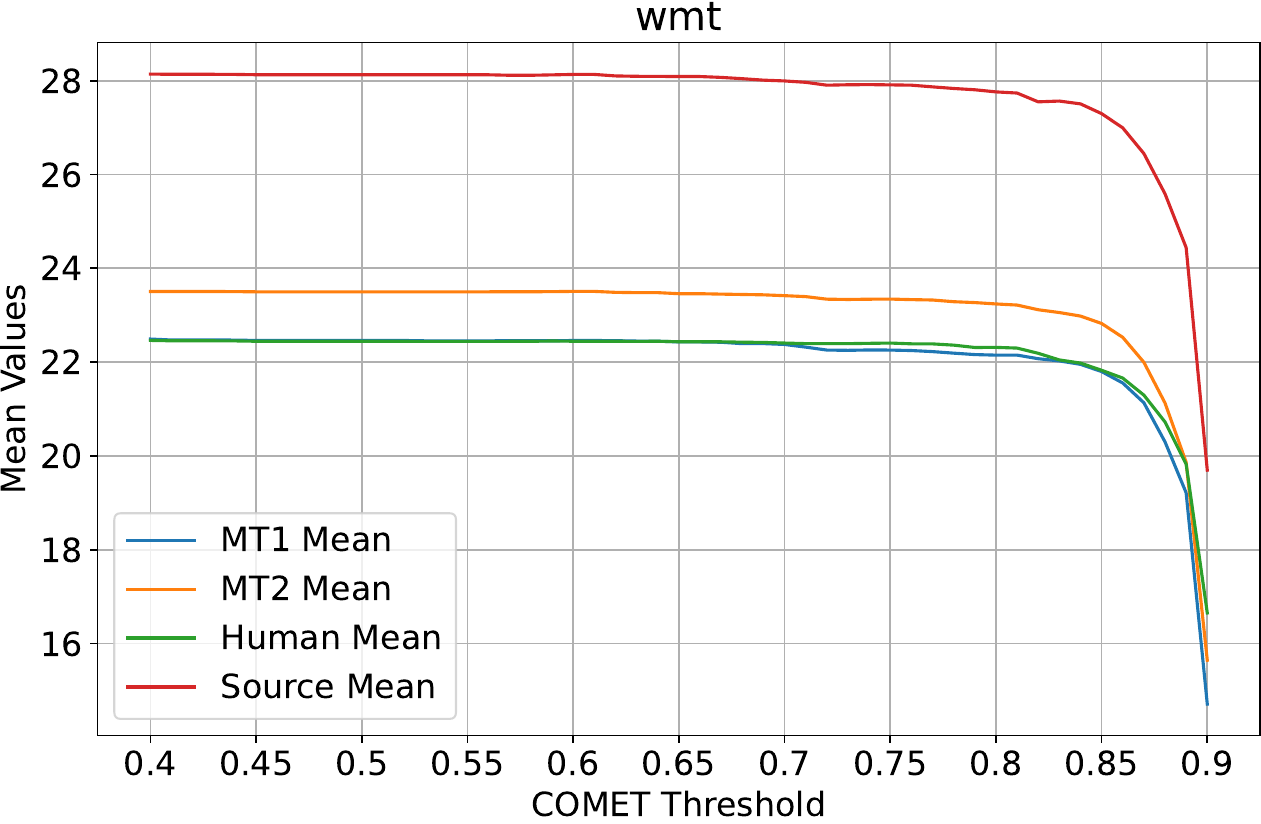} }}\quad%
    \subfloat[\centering Wilde]{{\includegraphics[width=7.5cm]{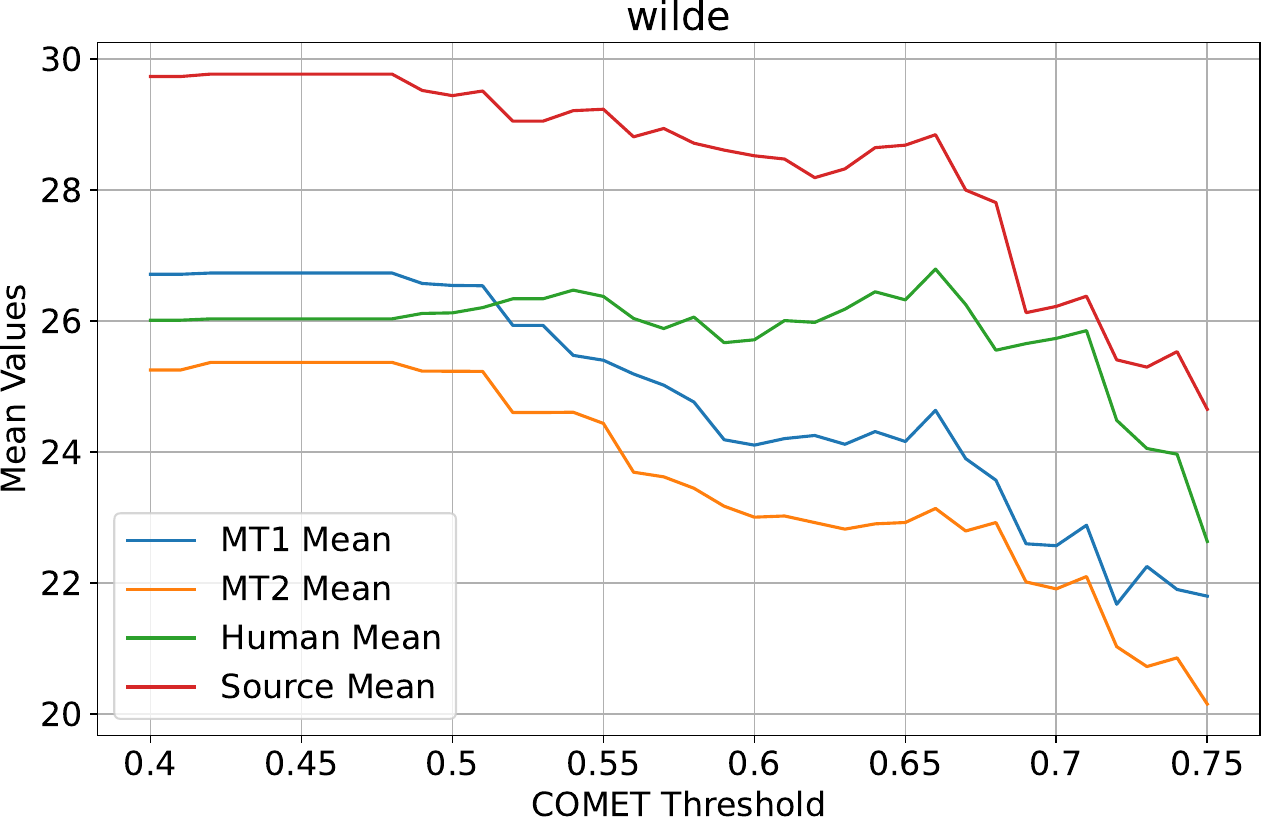} }}%
        \caption{Relationship between COMET scores of the MT and the $LV^2$ measure. As a proxy of translation quality, we use COMET score threshold to filter out low-quality translations. Higher values of $LV^2$ mean more diverse surprisal distribution. }%
    \label{fig:comet_lv2}%
\end{figure*}

We use the measures introduced in the previous section to confirm one of our initial hypotheses: 
\textit{Is NMT producing translations that are more uniform in terms of surprisal distribution than a human?}
This hypothesis posits that source sentences characterized by highly uneven surprisal distributions would exhibit more uniformity upon translation by MT systems, a phenomenon not expected to occur with human translations. 

We measured this across multiple datasets: For English to French, we use the Books corpus \cite{Zhu_2015_ICCV} (\textit{books}), Global Voices \cite{nguyen-daume-iii-2019-global} (\textit{global}), Newstest2014 \cite{bojar-etal-2014-findings} (\textit{wmt}), and a poem by Oscar Wilde translated into French by Jean Guiloineau (\textit{wilde}). 
For experiments in the English-Czech direction, we draw upon the dataset provided by \citet{zouhar2024quality,zouhar2023evaluating} (\textit{ORT}).

\begin{table}[!htp]\centering
\scriptsize
\begin{tabular}{llrrr}\toprule
\textbf{dataset} &\textbf{measure} &\textbf{HT} &\textbf{MT1} &\textbf{MT2} \\\midrule
\multirow{3}{*}{books} &$LV^2$ &0.39 &0.58 &0.42 \\
&$CV$ &0.42 &0.51 &0.50 \\
&$GV^2$ &0.46 &0.69 &0.54 \\ \midrule
\multirow{3}{*}{wmt} &$LV^2$ &0.43 &0.54 &0.58 \\
&$CV$ &0.49 &0.55 &0.57 \\
&$GV^2$ &0.46 &0.57 &0.64 \\ \midrule
\multirow{3}{*}{global} &$LV^2$ &0.69 &0.73 &0.78 \\
&$CV$ &0.65 &0.70 &0.63 \\
&$GV^2$ &0.72 &0.74 &0.80 \\ \midrule
\multirow{3}{*}{global\_doc} &$LV^2$ &0.72 &0.79 &0.83 \\
&$CV$ &0.68 &0.81 &0.82 \\
&$GV^2$ &0.76 &0.83 &0.86 \\ \midrule
\multirow{3}{*}{wilde} &$LV^2$ &0.16 &0.40 &0.53 \\
&$CV$ &0.07 &0.39 &0.54 \\
&$GV^2$ &0.16 &0.40 &0.53 \\ \midrule
\end{tabular}
\caption{Pearson's' $r$ for sentence-level surprisal uniformity of measurements between source and either HT, MT1 or MT2.}\label{tab:enfr_corr}
\end{table}


Table \ref{tab:enfr_corr} reveals that machine translations (MT\textit{x}) exhibit a better correlation with the source text's surprisal distribution than human translations (HT) for all measured indices: $LV^2$ (local variance squared), $CV$ (coefficient of variation), and $GV^2$ (global variance squared). Considering that human translators might spread surprisal over larger text units, we extended our analysis to document level in the \textit{global\_doc} dataset, treating each document as a single sequence of tokens for the purposes of surprisal estimation. Yet, the results did not support our hypothesis. 

However, the absolute values of the uniformity measures also indicated that MT is generally (with some exceptions, depending on the measure and the dataset) as uniform or less uniform than HT. This contradicts our initial hypothesis that MT will be more uniform in surprisal.
We hypothesized that this discrepancy might stem from errors in MT: if the MT system translates the input with some obvious mistakes, then these mistakes might be very surprising given the rest of the sentence. We used reference-free COMET (\texttt{wmt22-cometkiwi-da} \citealp{rei-etal-2022-cometkiwi}) scores to estimate the translation quality of the MT.

Figure \ref{fig:comet_lv2} illustrates the $LV^2$ measure's trends for instances where the machine translation (MT) COMET score surpasses a certain threshold (displayed on the $x$-axis). Uniformity between HT (green) and MT (two systems, blue and orange) remains steady across datasets, except for the \textit{wilde} dataset, which exhibits greater HT unevenness in high-scoring translations. This observation could imply that MT achieves higher surprisal uniformity than HT in highly creative content like poetry, when inaccurately translated segments are excluded from the evaluation. 

\begin{figure}%
\includegraphics[width=0.99\linewidth]{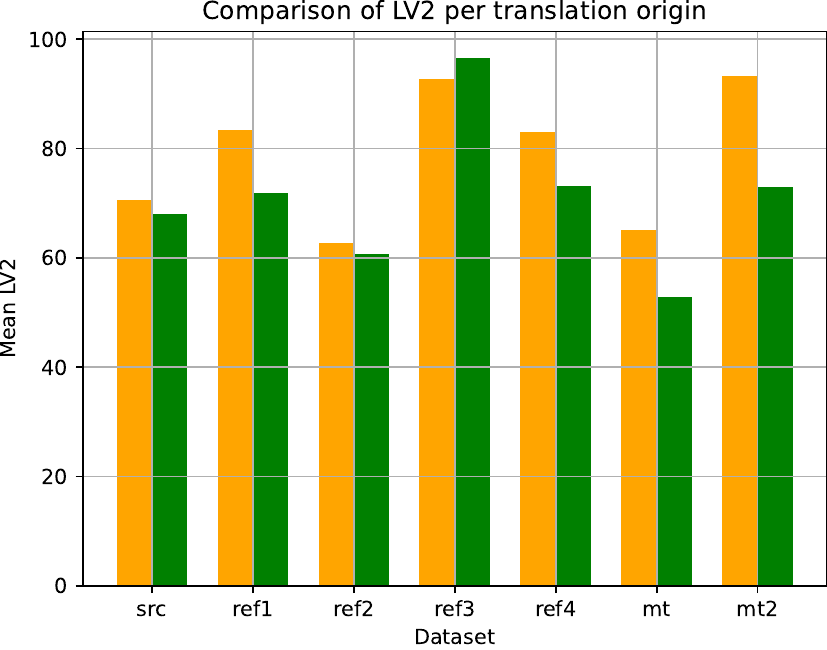}%
    \caption{Difference of $LV^2$ scores between all (orange) and high-quality (green) MT translations. Higher scores indicate less uniform text. }%
    \label{fig:all_hq_ort}%
\end{figure}


We have also experimented with another dataset, \textit{ORT}, which contains an English source sentence and four high-quality Czech translations. We compared surprisal distribution uniformity between the hypotheses and two MT engines.

The results are presented in Figure \ref{fig:all_hq_ort}.
We see that \texttt{MT1} usually scores as the most uniform, while \texttt{MT2} is among the least uniform translations, showing large variance among different MT systems. The results, again, contradict our initial hypothesis, that MT is inherently more uniform than HT -- it depends on both the human translator and MT system used. 
We again set a COMET threshold to filter out examples with low-quality MT.  We see that surprisal diversity is lower in high-quality translations (green), especially for \texttt{MT2}, and the third human reference (\texttt{ref3}) becomes the most diverse. \texttt{MT2}  is as diverse as the human references, with \texttt{ref3} being an exception.
Overall, we do not have reliable proof that MT produces texts that are more uniform in surprisal distribution than humans yet. Either our hypothesis is false, or our measurement methodology is flawed. One possible reason could be that the LMs we used to estimate the surprisals are trained on human text, not on MT outputs so it overestimates surprisal of some phenomena in MT. We plan further experiments to improve our methodology and extend the analysis to more datasets.


\subsection{O3: Alternative decoding algorithms}
\label{ssec:decoding}
Our proposed alternative to the beam search is based on sampling, MBR decoding, external resources like dictionaries, and a genetic algorithm. This method involves applying common GA operations -- mutation and crossover -- to a set of translation hypotheses, guided by a fitness function based on one or more MT metrics. We have presented this approach at ACL 2023 \citep{jon-bojar-2023-breeding} and we used it successfully in two applications:
\begin{itemize}
    \item Improving translation quality \citep{jon-etal-2023-cuni} (winning constrained system of WMT23 English-Czech and Czech-Ukrainian)
    \item Creating adversarial examples for MT evaluation metrics, we published sample datasets along with code to create similar datasets for arbitrary metric \citep{jon2024GAATME}
\end{itemize}
\begin{figure*}[ht]
\centering
\includegraphics[width=0.87\linewidth]{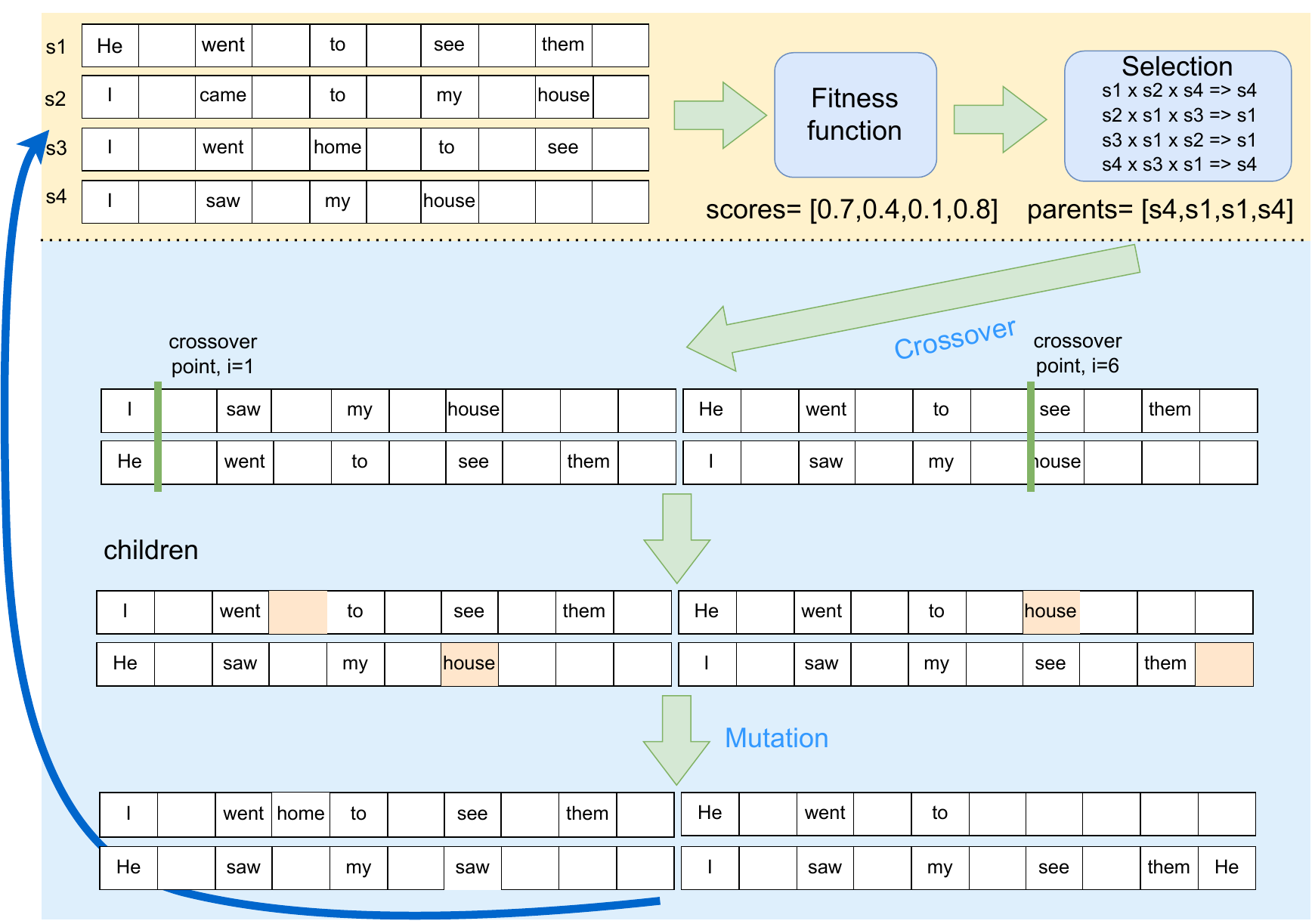}
\caption{One iteration of the GA algorithm for a population of 4 individuals. The steps with the yellow background are equivalent to simple reranking, the steps with the blue background introduce the operations of the genetic algorithm. Figure from \citet{jon-bojar-2023-breeding}.}
\label{fig:ga}
\end{figure*}

Illustration of the algorithm is presented in Figure \ref{fig:ga}. First, an initial set of candidate sentences is produced by an NMT model (e.g. via sampling or beam search). Candidates are then evaluated using a \textit{fitness function}: a weighted sum of scores from various MT metrics, possibly using MBR decoding with the initial translation candidates as pseudo-references. The scores are used by the \textit{selection method}, usually with some level of stochasticity (e.g. tournament selection), to choose the subset of translation candidates for as the parents for the next iteration of the genetic algorithm. 

The selected parents undergo a \textit{crossover} operation, where two translation hypotheses are split at a random word index and their segments swapped. Subsequently, the \textit{mutation} operation randomly alters the candidates by removing, adding, or replacing tokens, drawing new token choices from the set of words in the initial candidates, dictionary, or  the target language wordlist. These candidates are then scored again by the fitness function and the whole process is repeated. After a given number of iterations, the best-scoring candidate is selected as the final translation. The final translation always scores better or at least as well as the best translation from the initial candidates in the optimized metric.

Our findings indicate that employing a single MT metric in the fitness function tends to lead to overfitting, producing translations that achieve high scores according to that metric but contain serious translation errors. To explore this behavior, we set aside a designated \textit{held-out metric}, which is not used during the GA phase, to evaluate the translations generated by the algorithm. The results of this experiment can be interpreted as a rudimentary indicator of the robustness of the metric used in the fitness function: if the held-out metric scores are lower after GA than the initial hypotheses' scores,  i.e. if the GA process decreases the held-out scores, the metric is susceptible to adversarial examples and overfitting. In Table \ref{tab:adv_counts}, we show for how many instances within our test set this decrease in held-out metric scores occurs. CMT20 an QE20 refer to\textit{ wmt20-comet-da} and \textit{wmt20-comet-qe-da}. These results indicate the ease with which metrics can be misled to favor adversarially crafted translations that exploit their weaknesses, biases, and blind spots.

In \citet{jon2024GAATME}, we exploit this insight to create adversarial test sets for specific metrics, with a slight modification of the approach. Instead of using the held-out metric ex-post, to find the examples where the scores are worse, the held-out metric is directly incorporated into the fitness function, with a small negative weight. This change was made to actively steer the process toward creating adversarial translations. We select random words for possible mutations from an English wordlist. See Table \ref{tab:examples} in the appendix for examples of such adversarial translations for multiple metrics. Neural metrics show problems that are typical for them, like insensitivity to changes in named entities (probably due to embedding similarity of rare named entities) \citealp{amrhein-sennrich-2022-identifying}. BLEURT specifically seems to overwhelmingly prefer words that do not exist at all in the target language (they are part of the rare noise in the English wordlist).

\begin{table}[!htp]\centering
\scriptsize
\begin{tabular}{lcc}\toprule
$O$ &$O_{init}+m_o<O_{ga}$ & ... $ \land H_{init}>H_{ga}+m_h$ \\\midrule
CMT20 &128 (85\%) &57 (38\%) \\
QE20 & 148 (99\%) & 142 (95\%) \\
BLEU &150 (100\%) & 113 (75\%) \\
\bottomrule
\end{tabular}
\caption{Number of examples which improved in optimization metric after GA (2nd column) and at the same time deteriorated in held-out metric (3rd column). From \citet{jon-bojar-2023-breeding}.} \label{tab:adv_counts}
\end{table}

\begin{table}[!h]\centering

\scriptsize
\begin{tabular}{lccc}\toprule
Fitness  & +   & - & = \\\midrule
BLEU  & 22\%/1\%&29\%/7\%&49\%/92\% \\
CHRF   & 13\%/1\%&69\%/65\%&18\%/33\%  \\
CMT20 & \textbf{54\%}/23\% &\textbf{ 39\%}/32\% &\textbf{7\%}/45\% \\
CMT20+QE20+BLEU & \textbf{62\%}/\textbf{43\%} & \textbf{35\%}/\textbf{35\%} & \textbf{3\%}/\textbf{23\% }\\
\bottomrule
\end{tabular}
\caption{Percentage of examples where the held-out score (UniTE) improves (+), degrades (-), or doesn't change (=) for GA compared to best log-prob (before slash) or MBR reranking (after slash). Bold results are those where the held-out scores improve for more examples than deteriorate. From \citet{jon-bojar-2023-breeding}.} \label{tab:ratio}
\end{table}

On the other hand, we have found that the robustness can be improved by combining multiple metrics in the fitness function (we use a weighted sum of the values). In Table \ref{tab:ratio} we compare post-GA translations with the best (in terms of model log-prob) initial MT translation and the best translation after MBR reranking with the same metrics (i.e. after reranking with the fitness function used in the GA). We see that for the combination of CMT20, QE20, and BLEU beats or draws with the MBR reranking in most of the examples, while the standalone metrics (e.g. only CHRF) are performing worse, harming more examples than they improve. An illustration of behavior of fitness and held-out scores during GA in both the positive case (improved held-out score) and negative case (decreased held-out score) is shown in Figure \ref{fig:behavior} in the appendix.

These results were further extended and confirmed in our WMT23 submission \cite{jon-etal-2023-cuni}, where we show that newer versions of the COMET metrics are more robust, but combining multiple metrics is still beneficial. We have used GA to modify the outputs of two other submissions in English to Czech and Czech to Ukrainian tracks. In both cases, the modified outputs were scored slightly better by the human evaluators, although the difference was not significant. We expect more improvement could be obtained by tuning the parameters (mutation or crossover rate, number of generations, ways to combine metrics in the fitness function, \dots), which we did not do due to the computationally expensive nature of the process.

\subsection{O4: Alternative training objectives}
\label{ssec:cpc}

For Objective 4, we aimed to enhance NMT's ability for long-range planning within translations. We incorporated the Contrastive Predictive Coding (CPC) objective \citep{oord2019representation}. This algorithm introduces a regularization component to the training loss, designed to align internal representations in a selected layer in the current timestep to the representations in future timesteps.
One of the strengths of CPC is its focus not on predicting the exact future word, but rather on aligning future representations of words. This partially mitigates the issues with the ambiguity of the future, since embeddings tend to be similar for synonyms that would otherwise be penalized by the strict, single-correct-token prediction.

Traditional loss functions like cross-entropy struggle with predicting such high-dimensional targets like the internal representations directly. Instead, we focus on maintaining as much mutual information as possible between these representations.

The concrete implementation is through an InfoNCE loss, which maximizes the mutual information between an internal representation (in practice, we used embeddings in the last layer of the Transformer model) in the current step $c_t$ and the future step $c_{t+k}$ (positive example) and minimize it with respect to randomly selected representations from the same batch $c_j$ (negative examples):
$$
\mathcal{L}_{\mathrm{N}}=-\underset{C}{\mathbb{E}}\left[\log \frac{f_k\left(c_{t+k}, c_t\right)}{\sum_{c_j \in C} f_k\left(c_j, c_t\right)}\right]
$$

In practice, the function $f_k$ can be implemented in many ways, we have used a log-bilinear model: 
$$
f_k\left(c_{t+k}, c_t\right)=\exp \left(c_{t+k}^T W_k c_t\right)
$$
By minimizing this loss function, the mutual information between the representations is most preserved (derived in \citealp{oord2019representation}). Note that for simplicity, we are using single token representation $c_{t+k}$ as the target, but in practice, any embedding can be used, for example, a pooled, averaged embedding of multiple words, sentences, or paragraphs.

One implementational problem is that in the Transformer architecture, loss values and gradients for the backward pass are usually computed in one step for the whole batch, with possible relationships between the timesteps. This leads to a trivial solution to optimizing the loss function: instead of changing the past representations to contain more information about the future, the model will learn to store information about past representations in future ones. We solve this issue by using a frozen model to compute the future representations so that they are not taken into account for gradient computations.

We have carried out initial experiments with this implementation on MT. Overall, we have not seen improvements in automated MT quality scores yet, but we have only scratched the surface of potential experimental settings.






\subsection{Conclusions so far}
We have confirmed that surprisal estimates from language models are predictive of reading times on word level. On the sentence level, the discussion on the nature of the relationship between word surprisal distribution and sentence reading times and acceptability ratings is more complex. We have evaluated multiple surprisal distribution uniformity measures and our results show a small preference towards more uniformly distributed surprisals in human acceptability assessment.  We used these measures to test our hypothesis that MT produces more uniform translations than a human. Again, the results are mixed and vary greatly with chosen experimental settings (language model, tokenization, normalization, dataset). At best, we can see some evidence for our hypothesis in diverse text types, like books and poetry, but more research is necessary.

To replace the potentially problematic beam search, we have developed a novel decoding algorithm, based on sampling, Minimum Bayes Risk decoding, dictionaries, and a genetic algorithm. We demonstrated its effectiveness in 1) improving translation quality and 2) creating adversarial examples for arbitrary MT evaluation metrics. 

We have implemented an auxiliary training objective, Contrastive Predictive Coding, which aligns internal representations of NMT in each step with representations of future steps, as well as representations of more abstract text units.  We plan more experiments with this architecture.

\section{Future plans}
We have taken the first steps to tackle objectives \textbf{O1} through \textbf{O4}, defined in the Introduction. 

\subsection{O1 and O2: Identification of problematic texts and the root causes}
So far, we have only analyzed the surprisal-related properties of the text on the levels of words and sentences. Our results are not conclusive. We plan to run experiments on more diverse datasets, mostly from the literary domain. We also hypothesize that the the fact that the LMs are trained on natural text and we use surprisal estimates from them to evaluate uniformity on both natural and MT-produced text might affect the outcomes. We will train a new LM on a 50:50 mix of both types of texts.

 We will also evaluate alternative approaches to obtaining the word level (pseudo-)surprisal estimates, in place of language models. One possibility is to create a dataset of texts with binary sentence-level labels saying whether people consider the sentence surprising. We will train a classification model and use attribution methods \citep{javorský2023assessing} to calculate the influence of single words in the decision. 

We will move to longer textual units. We assume that on the level of a whole discourse, for example, a book or an article, some rhythmic changes between a quick and a slow pace of transmitting information, play a role in reader engagement and enjoyment of the text. We are in the process of obtaining/creating datasets from podcasts with metadata about where the user stopped listening.

We plan experiments that include a time dimension as well (i.e. timestamps of the texts), to see if the properties of the text on the Internet are already changing, due to the use of modern language tools. 

Overall, we did not find conclusive proof of MT producing more uniformly surprising texts yet. Even if the future planned experiments end inconclusively, the objectives \textbf{O3} and \textbf{O4}, which encompass most of the future work, can stand on their own. The proposed novel decoding algorithms and training objectives could improve other aspects of language processing and generation as well.

\subsection{O3: Decoding algorithms}
So far, we have only considered problems brought by the beam search decoding and we have not explored sampling algorithms used in current LLMs. Figuring out their influence on the properties of the text generated by LLMs will be the main part of the remaining work in this objective.

\subsection{O4: Training objectives}
The main focus of our future work will be on global training objectives. Such objectives could potentially improve the overall quality of natural language generation, aside from addressing the goals of our work. Thus, even if we fail to identify concrete issues in \textbf{O2}, developing these objectives can stand on its own.
Global planning is linked to intrinsic uncertainty in MT, i.e. the notion that a single source sentence has multiple possible translations. This fact is not modeled in the current objective of the MT -- each source example has only one target sentence probability distribution. This problem is more pronounced in globally operating training objectives since the uncertainty rises with increasing the distance from the prefix generated so far. 

Our CPC objective addresses this by predicting future internal representation instead of a single token. However, even the representations used currently are only single-point estimates in the multidimensional space of the model. We will look into incorporating the uncertainty in these embeddings as well, similar to \citet{Kesiraju_2020}, which could help to address the inherent ambiguity in generating translations. We will also conduct other experiments with CPC to verify its generalization and planning capabilities.

We will explore the use of segment-level training objectives and the effects of teacher forcing and its possible alternatives. We evaluate the ability of the alternative objectives to follow multiple possible paths in the translation using a multi-reference dataset \citep{manyref}.

In the later stages, we plan to focus on LLMs with their specifics, like RLHF (which can also be considered a segment-level objective). 

We might also move even further from traditional language modeling. An interesting approach to increase diversity and do away with the ``single correct past'' problem of auto-regressive, teacher-forced architectures is the use of diffusion models \citep{singh2023codefusion,li2022diffusionlm,lin2023text}.








\subsection{O5: Real world use-cases}

The more specific use cases (aside from saving the world from uniformity by fixing NLP) are closely related to the types of text we identify as problematic. So far, given both our intuition and experimental results in Section \ref{ssec:uid_mt}, literary translation seems like a promising testbed for our approach.


\bibliography{anthology,anthology_p2,custom}
\bibliographystyle{acl_natbib}

\appendix

\section{Appendix}
\label{sec:appendix}

\begin{table*}[!htpb]\centering
\scriptsize
\begin{tabular}{l@{\hskip5pt}p{0.23\textwidth}p{0.23\textwidth}p{0.23\textwidth}@{\hskip2pt}r@{\hskip3pt}r}
\textbf{Metric} &\textbf{MT} &\textbf{post-GA} &\textbf{Ref} &\textbf{MT score} &\textbf{GA score} \\\midrule
\multirow{6}{*}{\textbf{CMT22-QE}} &In the NHL, "France" caught 36 \textbf{games}, its \textbf{save} rate at 92.3\%. &In \textbf{yn}, "Frederic" clocked up 36 \textbf{ordain}, with \textbf{touchdown} rate at 92.3 \textbf{Basilica} &He has played 36 games in the NHL, where his save percentage is 92.3\%. &0.6407 &0.7679 \\
&The 31-year-old full-back will be on the \textbf{scoresheet} and could soon be in goal. &The fullback will \textbf{toilette on rotation }and could get into goal soon \textbf{fungo} &The thirty-one-year-old Pilsen native will be on the bench and could soon be in goal. &0.6901 &0.7242 \\ 
\midrule
\multirow{9}{*}{\textbf{CMT22}} &The highest ranked in the affair is \textbf{Berbr}, who no longer features in any of the football functions. &The highest profile in \textbf{glave} affair is \textbf{Piute Denten} who no longer longer figures \textbf{stanno} any football functions &The most senior figure in the affair is Berbr, who is no longer involved in any football function. &0.7947 &0.8104 \\
&Prince \textbf{William, Duke of Cambridge,} is wearing the same as Princes \textbf{George} and Louis shorts and a collared T-shirt. &Prince\textbf{ Pippo, Duke of Goldwyn}, dressed the same as Princes \textbf{Alexander} and Louis in shorts and a T-shirt &Prince William, Duke of Cambridge, and Princes George and Louis are wearing shorts and a polo shirt. &0.7675 &0.8402 \\ \midrule
\multirow{9}{*}{\textbf{CHRF}} &Interior has got respirators \textbf{significantly} cheaper than the Department of Health &Interior got respirators \textbf{mushy Asch} cheaper than the Ministry \textbf{oie} Ministry of Health\textbf{ natl . fur . LADT Goethe} &The Ministry of the Interior got respirators much cheaper than the Ministry of Health &0.5342 &0.8168 \\
&PVO: medium-cold war, outdated; short range - good, modern, relatively good number. &\textbf{unpaint:} medium-cold war, obsolete; short range - good, modern, relatively \textbf{Orth . enterable }number\textbf{. fugitively favorer POS SMDF R.A.A.F. pm . SM} &SHORAD: medium - cold war, obsolete; short range - good, modern, relatively favorable number. &67.9 &82.0 \\ \midrule
\multirow{5}{*}{\textbf{BLEU}} &\textbf{The picture}, which will serve as a Christmas card, was also posted by heir to the throne Prince Charles and wife Camilla. &	\textbf{knotty-leaved}, which will be \textbf{fat-shunning weeny-bopper} Christmas card, was also posted by the heir to the throne, Prince Charles, \textbf{dichlorodiphenyltrichloroethane duodenocholecystostomy cock-a-doodle-doos} his wife, Camilla. & The image, which will be used for the Christmas card, was also posted by the heir to the throne, Prince Charles, and his wife, Camilla.	& 34.3 & 68.6\\
\midrule
\multirow{9}{*}{\textbf{BLEURT}}  
& \textbf{By the time I} got off\textbf{ my seat, it was }gone.	& \textbf{Idun epicanthi} got \textbf{ achenium tundun terebinthial} off \textbf{Ladakhi Morgenthaler }gone\textbf{.. ecliptically scholium mesonasal}	& By the time I got off the deer-stand, he was gone. & 0.3965	& 0.8183 \\

& \textbf{His return to goal in the} NHL \textbf{eventually extended to more than two} months.	& \textbf{succinimid Badajoz hootchie-kootchie cheongsam}  NHL \textbf{taotai meromyarian Abyla Nadean vainer tenson} months 	& In the end, his time away from the NHL was extended by more than two months. &	0.5695	 &0.8510 \\

\bottomrule
\end{tabular}
\caption{Examples from the adversarial test set. Superfluous words in the post-GA translation and words from before GA that are missing post-GA are in bold. From \citet{jon2024GAATME}.}\label{tab:examples}

\end{table*}

\begin{figure}[h]
\centering
\begin{subfigure}[b]{1.00\linewidth}
\caption{}
\includegraphics[width=1.00\linewidth]{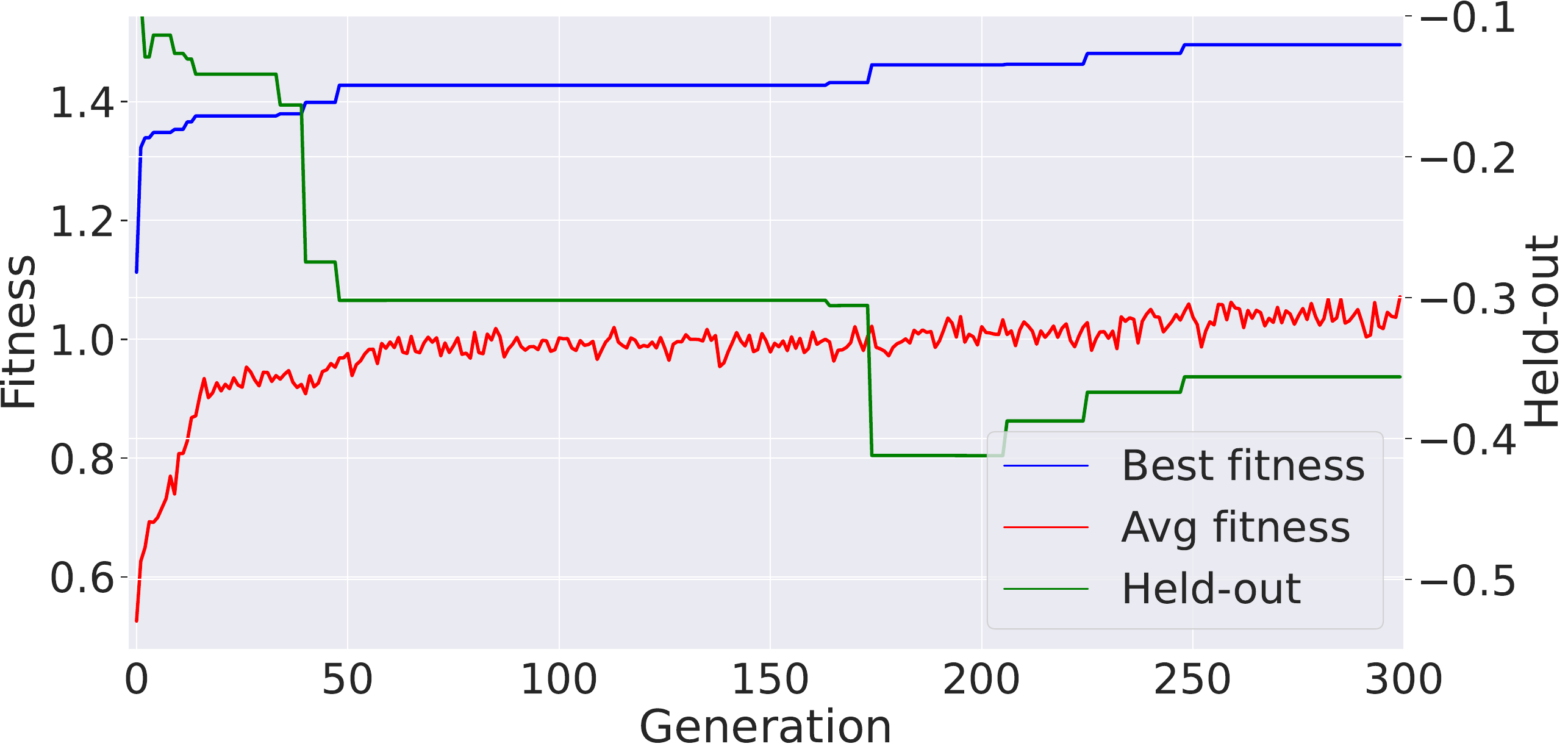}
\end{subfigure}

\begin{subfigure}[b]{1.00\linewidth}
\caption{}
\includegraphics[width=1.00\linewidth]{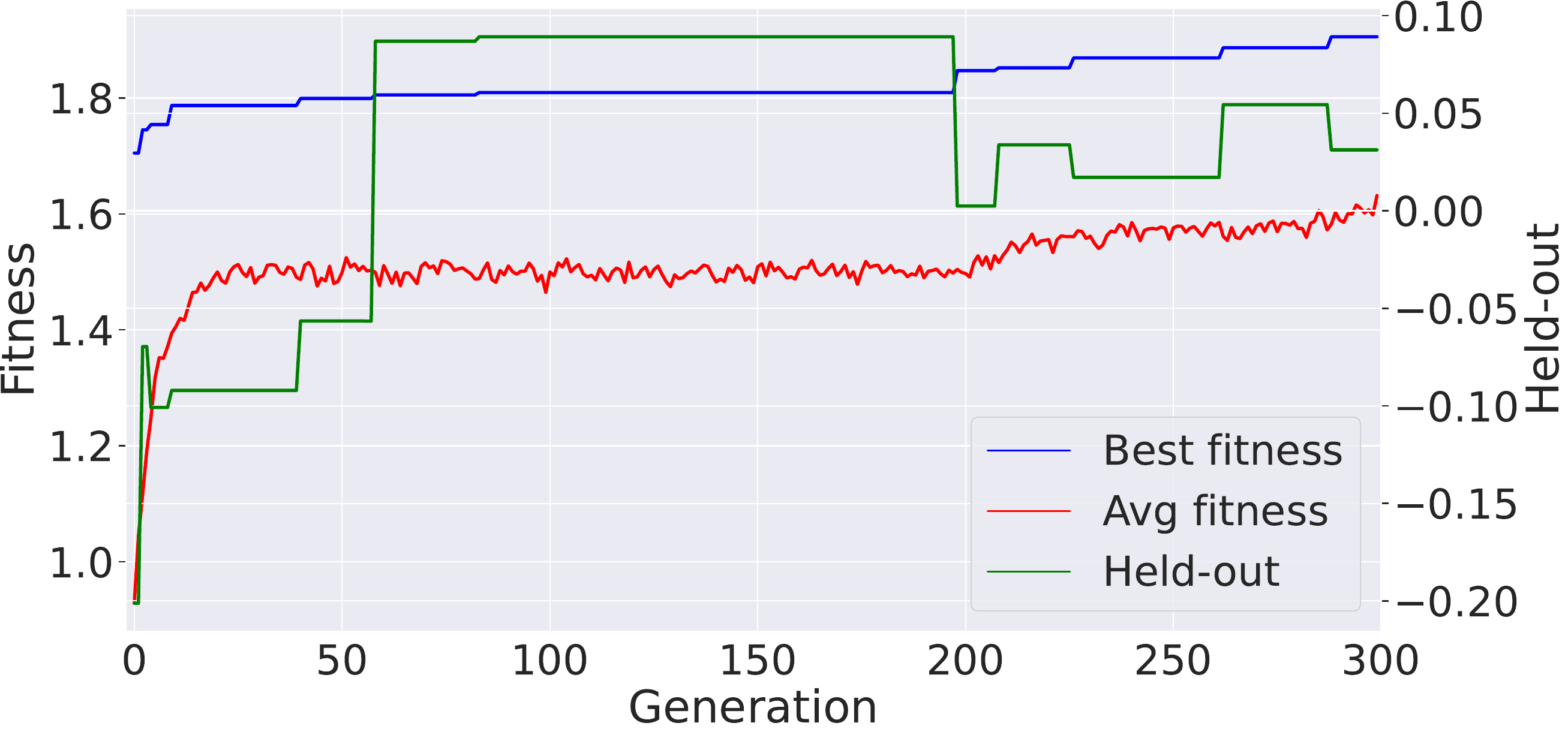}
\end{subfigure}
\caption{Behavior of best and population-average fitness, compared to held-out metric score $H$ of the best solution during a run of GA for two selected examples. Held-out score $H$ (UniTE) does not correlate well with the fitness metric (CMT20+QE+BLEU) and the GA is detrimental from the point of view of $H$ in Example a). In Example b), $H$ behaves similarly to the fitness function and the final held-out score is better than that of the best initial candidate. From \citet{jon-bojar-2023-breeding}.}
\label{fig:behavior}
\end{figure}

\end{document}